\documentclass[lettersize,journal]{IEEEtran}
\usepackage{amsmath,amsfonts}
\usepackage{algorithmic}
\usepackage{algorithm}
\usepackage{array}
\usepackage[caption=false,font=normalsize,labelfont=sf,textfont=sf]{subfig}
\usepackage{textcomp}
\usepackage{stfloats}
\usepackage{url}
\usepackage{verbatim}
\usepackage{graphicx}
\usepackage{cite}
\usepackage{hyperref}
\usepackage{float}
\usepackage{multirow}
\usepackage[flushleft]{threeparttable}
\usepackage{cleveref}
\usepackage{soul}
\usepackage{booktabs}
\usepackage{amsmath}
\usepackage{amssymb}
\usepackage{mathtools,euscript,yfonts,latexsym}
\usepackage{bm} 
\usepackage{color, xcolor} 
\definecolor{deepblue}{RGB}{0,0,0}
\definecolor{deepbluex}{RGB}{0,0,0} 

\begin{document}

\title{Mask Approximation Net: A Novel Diffusion Model Approach for Remote Sensing Change Captioning}

\author{Dongwei Sun, Jing Yao,~\IEEEmembership{Member,~IEEE}, Wu Xue, Changsheng Zhou, Pedram Ghamisi,~\IEEEmembership{Senior Member},~IEEE, Xiangyong Cao,~\IEEEmembership{Member,~IEEE}
\thanks{This work was supported by the Key R\&D Program of Zhejiang under Grant 2025C01075, the National Natural Science Foundation of China under Grants 62272375, 62201553, 12201145, 62272374, 62192781, 62250009, 62137002 and 62276208, Natural Science Foundation of Shaanxi Province (No. 2024JC-
JCQN-62), Natural Science Basic Research Program of Shaanxi Province (2024JC-JCQN-02), Project of China Knowledge Center for Engineering Science and Technology, and Project of Chinese academy of engineering “The Online and Offline Mixed Educational Service System for ‘The Belt and Road’ Training in MOOC China”. 
(\emph{Corresponding authors: Jing Yao; Xiangyong Cao})}

\thanks{D. Sun and X. Cao are with the School of Computer Science and Technology and the Ministry of Education Key Lab for Intelligent Networks and Network Security, Xi’an Jiaotong University, Xi’an 710049, China (e-mail:sundongwei@outlook.com, caoxiangyong@xjtu.edu.cn).}
\thanks{J. Yao is with the Aerospace Information Research Institute, Chinese Academy of Sciences, Beijing 100094, China (e-mail: yaojing@aircas.ac.cn).}
\thanks{W. Xue is with the Space Engineering University, Beijing 101416, China (e-mail: xuewu\_81@126.com).}
\thanks{C. Zhou is with the School of Mathematics and Statistics, Guangdong
University of Technology, Guangzhou, China (e-mail: chsh\_zh@gdut.edu.cn).
}
\thanks{P. Ghamisi is with the Helmholtz-Zentrum Dresden-Rossendorf, 09599 Freiberg, Germany, and also with the Lancaster Environment Centre, Lancaster University, LA1 4YR Lancaster, U.K. (e-mail: p.ghamisi@hzdr.de).}
}



\maketitle

\begin{abstract}
Remote sensing image change description represents an innovative multimodal task within the realm of remote sensing processing. This task not only facilitates the detection of alterations in surface conditions, but also provides comprehensive descriptions of these changes, thereby improving human interpretability and interactivity.
Current deep learning methods typically adopt a three-stage framework consisting of feature extraction, feature fusion, and change localization, followed by text generation. Most approaches focus heavily on designing complex network modules but lack solid theoretical guidance, relying instead on extensive empirical experimentation and iterative tuning of network components. This experience-driven design paradigm may lead to overfitting and design bottlenecks, thereby limiting the model’s generalizability and adaptability.
To address these limitations, this paper proposes a paradigm that shift towards data distribution learning using diffusion models, reinforced by frequency-domain noise filtering, to provide a theoretically motivated and practically effective solution to multimodal remote sensing change description.
The proposed method primarily includes a simple multi-scale change detection module, whose output features are subsequently refined by a well-designed diffusion model. 
Furthermore, we introduce a frequency-guided complex filter module to boost the model performance by managing high-frequency noise throughout the diffusion process. We validate the effectiveness of our proposed method across several datasets for remote sensing change detection and description, showcasing its superior performance compared to existing techniques. The code will be available at \href{https://github.com/sundongwei}{MaskApproxNet}. 
\end{abstract}


\section{Introduction}
\IEEEPARstart{R}{emote} sensing change detection is a vital task in analyzing and monitoring temporal variations on the Earth's surface through satellite or aerial imagery. By comparing images acquired at different time intervals, this task aims to identify changes such as land use modifications \cite{daudt2018fully}, deforestation, urbanization \cite{wen2019accurate}, and environmental degradation \cite{de2020change,khan2017forest}. \textcolor{deepbluex}{Traditionally, change detection   \cite{wang2024changeminds,yu2024maskcd,9870837,9736956} has relied on pixel-level or object-level approaches to highlight areas with significant differences between bi-temporal images \cite{10855580,10795250}, providing essential information for environmental monitoring, disaster management\cite{10423067,10716673,10750064}, and urban planning.
However, while these methods are effective at detecting the presence or absence of changes, they often lack the ability to describe the nature and significance of the detected changes in a human-readable format \cite{hosseinzadeh2021image,remoteclip}}. 
This limitation highlights the need for remote sensing change captioning.

\begin{figure}[t!] 
    \centering
    \includegraphics[width=0.5\textwidth]{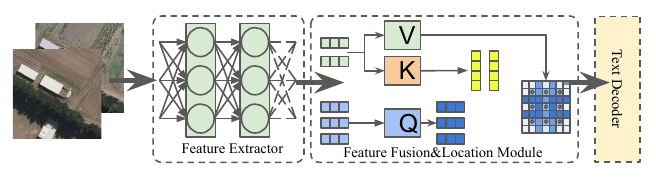} 
    \begin{minipage}{0.5\textwidth}
        \raggedright \scriptsize (a) Previous three-stage methods require feature extraction, feature fusion and localization, and a text decoder to generate the final change detection description. The entire process is cumbersome and heavily reliant on the network architecture design.
    \end{minipage}
    \hfill
    \includegraphics[width=0.5\textwidth]{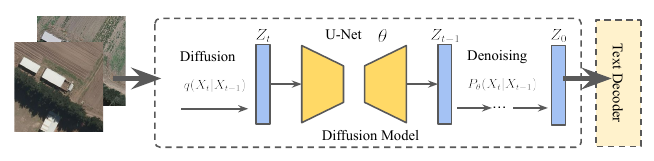} 
    \begin{minipage}{0.5\textwidth}
        \raggedright \scriptsize (b) The method proposed in this paper is simple and effective, avoiding heavy reliance on neural network design while effectively learning the data feature distribution.
    \end{minipage}
    \caption{A comparison between previous three-stage methods and the proposed method.} 
    \label{fig:figure0}
\end{figure}

\textcolor{deepbluex}{In recent years, remote sensing change captioning has garnered significant attention due to its capability to provide detailed, natural-language descriptions of temporal variations on the Earth's surface \cite{wang2025change,10504785}. Despite advances in this area, existing deep learning-based methods \cite{8240966,10679571,9745546} that are predominantly based on Convolutional Neural Networks (CNNs) exhibit several shortcomings, hindering their practical applicability {\cite{bruzzone2000automatic,qiu2021describing}}.} \textcolor{deepblue}{ A primary concern lies in their reliance on a three-stage framework—feature extraction, feature fusion, and change localization—followed by text generation, where the design of complex network modules is mainly driven by empirical experimentation rather than solid theoretical foundations, as illustrated in Figure \ref{fig:figure0} (a). This experience-driven approach often leads to overfitting and design bottlenecks, restricting the models’ robustness and adaptability. Consequently, their generalizability across diverse datasets and application scenarios remains limited, especially when faced with domains that differ significantly from the training data.However, diffusion models offer a promising alternative. As a likelihood-based generative approach, diffusion models employ a Markov chain to progressively add noise to data and learn to reconstruct samples by reversing this noising process. This stochastic process inherently captures data diversity and complexity, which effectively reduces overfitting \cite{jiang2024comat, luo2023semantic} and enhances model generalization. Unlike conventional CNN-based methods that are prone to overfitting and often struggle with robustness and reliability in real-world, dynamic remote sensing scenarios, diffusion models are better suited to handle the diverse and complex data distributions encountered in change captioning tasks, thereby improving practical applicability and performance.}

To address the above-mentioned limitations, in this article, we propose a novel method by leveraging diffusion models for remote sensing image change captioning \cite{liu2024diffusion}. As illustrated in Figure \ref{fig:figure0} (b), our approach diverges from previous methods that depend on complex attention networks to generate change masks. Instead, we utilize diffusion models, which prioritize learning data distributions over mere feature extraction. 
\textcolor{deepbluex}{Diffusion models have shown remarkable success in both continuous visual data \cite{gao2023implicit, ho2022classifierfreediffusionguidance} and discrete language data domains \cite{li2022diffusion, gong2022diffuseq}, producing high-quality samples while deeply capturing connections between pixels, words, and image-text relationships in multi-modal tasks \cite{zheng2023rdm,10689651}}. By incorporating a paradigm shift that enhances their ability to grasp temporal complexities, these models are particularly adept at handling the dynamic intricacies of remote sensing imagery.
Furthermore, we integrate a multi-scale change captioning module that identifies changes at various scales and refines the outputs through a diffusion process, resulting in more robust and adaptive caption generation. The proposed strategy not only simplifies the entire pipeline for change detection and captioning but also enhances generalizability and robustness by eliminating the need for intricate network designs. Besides, we introduce a frequency-guided complex filter module that effectively manages high-frequency noise, ensuring that the learned representations remain accurate and resilient, thereby improving the overall quality of the generated captions.

Specifically, our proposed approach provides several key advantages over existing methods. First, it introduces the use of diffusion models for the remote sensing image change captioning task, marking the first time this method has been applied in this context. This novel attempt not only improves performance but also simplifies the conventional pipeline on this task. Second, rather than merely focusing on image feature learning, we emphasize learning feature distributions, significantly improving the model's ability to capture complex and dynamic changes in remote sensing imagery. Additionally, we integrate a frequency-guided complex filter module into the proposed framework to effectively manage high-frequency noise, thereby preserving the integrity of the learned feature distributions. Finally, extensive experiments conducted on multiple remote sensing change captioning datasets demonstrate that our method consistently outperforms state-of-the-art approaches in terms of captioning accuracy and robustness. These advancements make our framework particularly well-suited for real-world applications where interpretability and adaptability are essential.

In summary, our work presents a robust and adaptable framework for remote sensing change captioning by integrating diffusion models and a frequency-guided filtering mechanism. The key highlights of our proposed method include the adaptation of diffusion models for change captioning, the transition of focus from simple feature learning to feature distribution learning, and the incorporation of a frequency-guided filtering mechanism to effectively manage high-frequency noise. This comprehensive approach ensures that our model not only performs well on standard datasets but also excels in challenging real-world scenarios. The contributions of this paper can be summarized as follows:

\begin{enumerate}
\item \textbf{Introduction of the Diffusion Model}: \textcolor{deepblue}{To the best of our knowledge, our work is the first to introduce diffusion models specifically into the image processing component of the remote sensing image change captioning task. By leveraging the strong theoretical foundation of diffusion models to effectively model complex temporal variations in multi-temporal remote sensing imagery, our approach aims to improve performance while simplifying the overall task framework.}
\item \textbf{Feature Distribution Learning}: By shifting focus from basic feature learning to learning feature distributions, we significantly improve the model's ability to capture complex, dynamic changes in remote sensing imagery. 
\item \textbf{Frequency-Guided Filtering Mechanism}: The integration of a frequency-guided complex filter module effectively manages high-frequency noise, preserving the integrity of learned feature distributions.
\item \textbf{Extensive Experimental Validation} Our method demonstrates superior performance over state-of-the-art approaches across multiple remote sensing change captioning datasets, showcasing its accuracy and robustness.
\end{enumerate} 

The paper is organized as follows: Section \ref{Related Work} provides a review of related work, while Section \ref{Method} outlines the proposed method in detail. Section \ref{Experimental} presents the experimental setup, including comprehensive descriptions and an analysis of the results. A brief summary is finally concluded in Section \ref{Conclusion}.

\section{Related Work} \label{Related Work}
\subsection{Change Captioning}
The \textcolor{deepblue}{Remote Sensing Image Change Captioning(RSICC)} task \cite{hoxha2024dubai} has garnered significant attention in recent years due to its capability to describe differences between bitemporal remote sensing (RS) images using natural language. Park et al. \cite{park2019robust} presents a novel Dual Dynamic Attention Model (DUDA) to perform robust change captioning. Hoxha et al. \cite{hoxha} introduced early and late feature fusion strategies to integrate bitemporal visual features, utilizing an RNN and a multiclass SVM decoder for generating change captions. Chouaf et al. \cite{chouaf} were among the first to explore the RSICC task, employing a CNN as a visual encoder to capture temporal scene changes and an RNN as a decoder to produce descriptive change captions.  

\begin{figure*}[t!] \centering
\includegraphics[width=\textwidth]{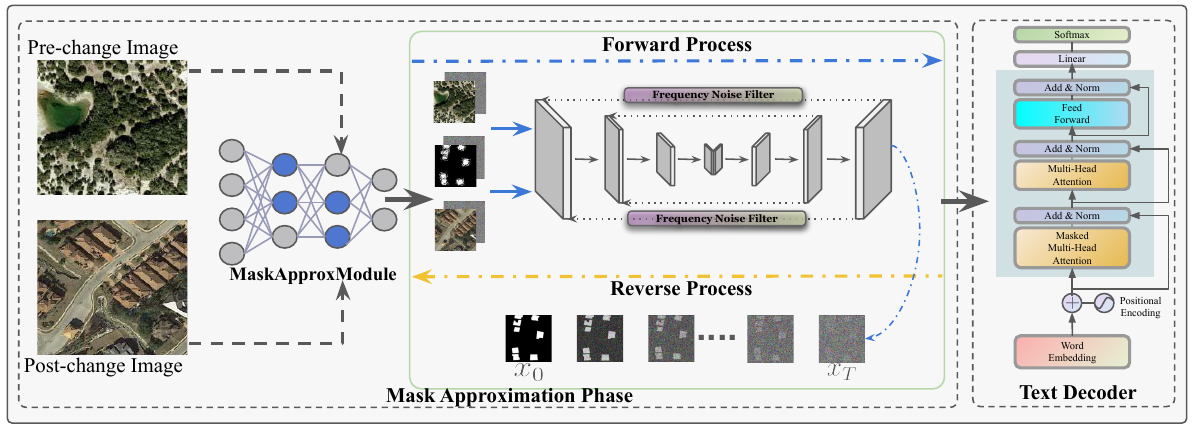}
    \caption{\textcolor{deepblue}{The overall framework of the proposed method includes two main stages. The first stage employs a conditional diffusion model that generates change detection (CD) maps by progressively denoising Gaussian noise, guided by the pre-change and post-change images ($I_{pre}$ and $I_{post}$). This iterative process allows effective modeling of complex temporal changes. The second stage decodes the generated CD maps into descriptive text, completing the remote sensing change captioning task.}} \label{fig:figure1}
\end{figure*}

Transformer networks \cite{NIPS2017_3f5ee243}, incorporating the multi-head attention (MHA) mechanism, have gained prominence in image analysis and achieved notable success in image change captioning. \textcolor{deepbluex}{Liu et al. have made a series of significant contributions to RSICC. They first proposed a transformer-based encoder-decoder framework featuring a dual-branch transformer encoder for scene change detection and a multistage fusion module to integrate multilayer features for generating change descriptions \cite{liuc}. Building on this, they introduced progressive difference perception transformer layers to better capture both high-level and low-level semantic changes \cite{liuc2}. More recently, they developed a prompt-based approach that leverages pretrained large language models (LLMs), using visual features, change classes, and language representations as prompts to guide a frozen LLM in generating change captions \cite{liuc3}. Kuckreja et al. \cite{kuckreja2024geochat} expanded the image-text pairs of existing different remote sensing datasets, a multifunctional model Geo Chat was proposed to generate change captions.}
Chang et al. \cite{changs} developed an attentive network for RS change captioning, named Chg2Cap, which harnesses the strengths of transformer models commonly used in NLP. Sun et al. \cite{sundw} proposes a lightweight Sparse Focus Transformer for remote sensing image change captioning, which significantly reduces parameters and computational complexity while maintaining competitive performance. In DiffusionRSCC \cite{yu2025diffusion} introduced an innovative diffusion model for RSICC, employing a forward noising and reverse denoising process to learn the probabilistic distribution of the input. Zhou et al.\cite{sen} proposed a bitemporal pretraining method that leverages self-supervised learning on a large-scale bitemporal RS image dataset. This approach reduces data distribution and input gaps, resulting in more suitable features for RSICC and improved model generalization.
\subsection{Generative Models}
Generative Adversarial Networks (GANs), as a typical generative approach, are widely used in the field of remote sensing imagery, particularly in change detection. The research in \cite{qiu2023dual} introduces a Dual Attentive Generative Adversarial Network (DAGAN) for high-resolution remote sensing image change detection. By designing a multi-level feature extractor and a multi-scale adaptive fusion module, DAGAN effectively integrates features at various levels, enhancing the accuracy of change detection.
Wang et al. \cite{wang2022cd} presents an unsupervised change detection method named CD-GAN, tailored for heterogeneous remote sensing images acquired from different sensors. By integrating generative adversarial networks, CD-GAN can detect change regions without the need for image registration, improving robustness and accuracy. Ren et al. \cite{ren2020unsupervised} proposes a GAN-based method that generates better-registered images, which mitigates the impact of misregistration on change detection and enhances detection performance.

Within the enormous models for
CD, \textcolor{deepblue}{Denoising Diffusion Probabilistic Model(DDPM)} based \cite{wen2024transc} architectures emerge with distinguished advantages
over traditional CNNs and transformers. In DDPM-CD \cite{bandara2022ddpmcd} by pre-training the DDPM on a large set of unlabeled remote sensing images, multi-scale feature representations are obtained. A lightweight change detection classifier is then fine-tuned to detect precise changes.
Wen et al. \cite{wen2023gcd} designed to guide the generation of change detection maps by exploiting multi-level difference features. The Similar ideas have also applied in medicine field. Yu et al. \cite{yu2025diffusion} employs a noise predictor conditioned on cross-modal features to generate human-like descriptions of semantic changes between bi-temporal remote sensing image pairs. Zhang et al. \cite{zhang2023graph} presents a graph attention-guided diffusion model tailored for liver vessel segmentation. By integrating graph attention mechanisms with diffusion probabilistic models, the approach effectively captures complex vascular structures in liver images.

\section{Methodology} \label{Method}
In this section, we propose a novel diffusion model-based approach for the remote sensing image change captioning task, leveraging mask generation for change captioning. The proposed method consists of two key phases: 1) the mask approximation phase, and 2) the text decoding phase. The primary objective of the first phase is to construct a mapping from the change distribution to the standard Gaussian distribution by utilizing the designed MaskApproxNet network, which extracts change features from the pre-change image and post-change image. In the reverse denoising process, the prior mask is transformed from the standard Gaussian distribution back to the real data distribution using a denoiser network.

In the following parts, we will first introduce the preliminaries of diffusion models using DDPM as an example. Then, we will detail the entire implementation process in the mask approximation phase, including the specific design of MaskApproxNet and the frequency noise filter. Finally, we will elaborate the text decoding process.

\begin{figure}[!t] \centering
    \includegraphics[width=0.48\textwidth]{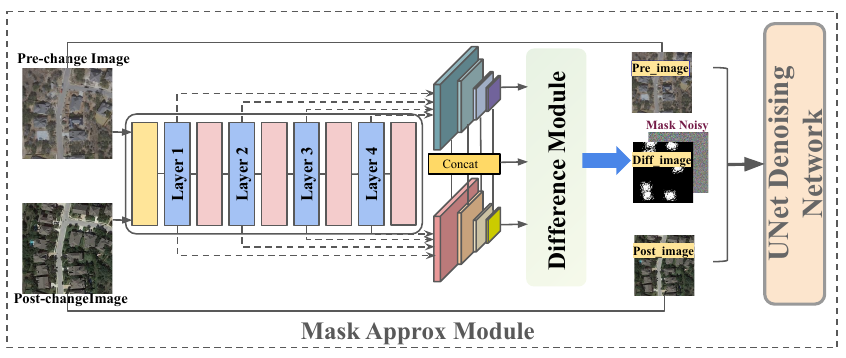}
    \caption{The flowchart highlights the Mask Approximation Phase, showcasing the process of multi-scale feature extraction, difference computation, and feature refinement to generate a precise difference image for change detection.} \label{fig:figure2}
\end{figure}

\subsection{Denoising Diffusion Models}
Diffusion models operate by defining a forward Markov process that gradually transforms data into noise and a reverse process that reconstructs data from noise. Formally, given an initial data distribution \(\mathbf{x}_0 \sim q(\mathbf{x}_0)\), the forward process generates a sequence of variables \(\mathbf{x}_1, \mathbf{x}_2, \ldots, \mathbf{x}_T\) using a transition kernel \(q(\mathbf{x}_t \mid \mathbf{x}_{t-1})\). By applying the Markov property and the chain rule, the joint distribution of the sequence conditioned on \(\mathbf{x}_0\) can be expressed as:  

\begin{equation}
\label{eq1}
q(\mathbf{x}_1, \ldots, \mathbf{x}_T \mid \mathbf{x}_0) = \prod_{t=1}^{T} q(\mathbf{x}_t \mid \mathbf{x}_{t-1}).  
\end{equation}

In the context of denoising diffusion probabilistic models (DDPMs), the transition kernel \(q(\mathbf{x}_t \mid \mathbf{x}_{t-1})\) is designed to gradually transform the data distribution \(q(\mathbf{x}_0)\) into a simple, tractable prior. A common choice for this kernel is a Gaussian perturbation, defined as:  

\begin{equation}
\label{eq2}
q(\mathbf{x}_t \mid \mathbf{x}_{t-1}) = \mathcal{N}(\mathbf{x}_t; \sqrt{1-\beta_t} \mathbf{x}_{t-1}, \beta_t \mathbf{I}),  
\end{equation}
where \(\beta_t \in (0, 1)\) is a predefined hyperparameter. This Gaussian kernel allows the joint distribution to be marginalized analytically, yielding:  

\begin{equation}
\label{eq3}
q(\mathbf{x}_t \mid \mathbf{x}_0) = \mathcal{N}(\mathbf{x}_t; \sqrt{\bar{\alpha}_t} \mathbf{x}_0, (1-\bar{\alpha}_t) \mathbf{I}),  
\end{equation}
where \(\alpha_t = 1 - \beta_t\) and \(\bar{\alpha}_t = \prod_{s=0}^t \alpha_s\). A sample \(\mathbf{x}_t\) can then be generated from \(\mathbf{x}_0\) using the transformation:  

\begin{equation}
\label{eq4}
\mathbf{x}_t = \sqrt{\bar{\alpha}_t} \mathbf{x}_0 + \sqrt{1-\bar{\alpha}_t} \boldsymbol{\epsilon},  
\end{equation}
where \(\boldsymbol{\epsilon} \sim \mathcal{N}(\mathbf{0}, \mathbf{I})\). When \(\bar{\alpha}_T \approx 0\), the final state \(\mathbf{x}_T\) approximates a Gaussian distribution, \(q(\mathbf{x}_T) \approx \mathcal{N}(\mathbf{0}, \mathbf{I})\).  

Intuitively, the forward process adds noise step by step, erasing data structure until only noise remains. To generate new data, DDPMs reverse this process. Starting with a noise sample \(\mathbf{x}_T \sim \mathcal{N}(\mathbf{0}, \mathbf{I})\), the reverse Markov chain iteratively removes noise to reconstruct the data. 

The reverse process is parameterized by a prior \(p(\mathbf{x}_T) = \mathcal{N}(\mathbf{0}, \mathbf{I})\) and a learnable transition kernel \(p_\theta(\mathbf{x}_{t-1} \mid \mathbf{x}_t)\), which is modeled as:  

\begin{equation}
\label{eq5}
p_\theta(\mathbf{x}_{t-1} \mid \mathbf{x}_t) = \mathcal{N}(\mathbf{x}_{t-1}; \mu_\theta(\mathbf{x}_t, t), \Sigma_\theta(\mathbf{x}_t, t)),  
\end{equation}
where \(\mu_\theta(\mathbf{x}_t, t)\) and \(\Sigma_\theta(\mathbf{x}_t, t)\) are neural network outputs parameterized by \(\theta\). The reverse chain starts from \(\mathbf{x}_T\) and iteratively samples \(\mathbf{x}_{t-1}\) until reaching \(\mathbf{x}_0\), yielding a generated data sample.  

\subsection{Mask Approximation Phase}
\textcolor{deepblue}{Existing methods usually rely on multi-stage architectures handling feature extraction, fusion, localization, and description separately. This leads to inefficiencies, as numerous stages cause misalignment between image features and text, propagating errors. Moreover, the complexity arises from empirically designed network modules lacking strong theoretical support, limiting model robustness and scalability.}

In contrast, our proposed algorithm model, shown in Figure \ref{fig:figure1}, simplifies the process by integrating these stages into a single, unified structure. The model is divided into two main phases, designed to process the given remote sensing images \(I_{pre}\) and \(I_{post}\) efficiently. The first stage, the \textbf{mask approximation phase}, handles simple feature processing and combines the results with a noisy mask, which is then passed through the UNet denoising network to generate change features. These features are subsequently forwarded to the second stage, the \textbf{text decoding}, to generate change detection descriptions.

By streamlining the entire process into a single flow, we eliminate the need for separate stages, thereby reducing the potential for error propagation and improving computational efficiency. This design choice allows for a more direct and effective approach to change detection and captioning, making our method not only simpler but also more robust and efficient. Next, we will introduce each of these two phases in detail.

\subsubsection{Mask Approx Module}
The proposed framework as illustrated Figure \ref{fig:figure2}, which processes a pair of bi-temporal remote sensing (RS) images, denoted as \( I_{\text{pre}} \) (pre-change image) and \( I_{\text{post}} \) (post-change image), through a systematic pipeline. First, a Siamese ResNet backbone is utilized to extract multi-scale features at four levels, represented as \( X_i^{\text{pre}} \) and \( X_i^{\text{post}} \) for \( i \in \{1, 2, 3, 4\} \), where:  
\begin{equation}
X_i^{\text{pre}} = \text{ResNet}(I_{\text{pre}}, i), \quad X_i^{\text{post}} = \text{ResNet}(I_{\text{post}}, i).
\end{equation}
These features are then input to a difference module, which separately processes the multi-scale features and combines them through a convolution layer to generate rich feature encodings, denoted as \( \overline{X} \):  
\begin{equation}
\overline{X} = \text{Conv}(\text{DifferenceModule}(X_i^{\text{pre}}, X_i^{\text{post}})).
\end{equation}
 
The encoded representations \( \overline{X} \) are passed to a decoder that upsamples the features to match the spatial resolution of the input images. The decoder employs two transpose convolution layers:  
\begin{equation}
\overline{X}_{\text{upsampled}} = \text{TransposeConv}(\overline{X}),
\end{equation}
followed by a residual convolutional block for feature enhancement:  
\begin{equation}
\overline{X}_{\text{refined}} = \text{ResidualBlock}(\overline{X}_{\text{upsampled}}).
\end{equation}
 
Finally, a convolution layer is applied to the refined features to produce the predicted difference image \( \text{DiffImage} \):  
\begin{equation}
\text{DiffImage} = \text{Conv}(\overline{X}_{\text{refined}}).
\end{equation}

\begin{figure}[t!] \centering
    \includegraphics[width=0.48\textwidth]{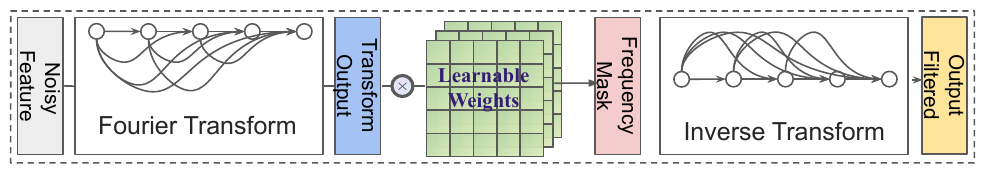}
    \caption{The flowchart illustrates the overall process of the Frequency-Guided Channel Fusion (FGCF) method, encompassing frequency-domain transformation, frequency-guided weighting, inverse transformation to the spatial domain, and channel-wise attention for feature fusion.} \label{fig:figure3}
\end{figure}

This integrated framework effectively extracts, processes, and decodes change representations to achieve accurate change detection.

To achieve denoising, we employ U-Net as the core network. During the forward process, the change label \(\mathbf{x}_0\) undergoes a sequence of \(T\) steps where Gaussian noise is progressively added. Conversely, the reverse process is designed to reconstruct the original data by systematically removing the noise. The reverse process is mathematically defined as, 

\begin{equation}
p_\theta(\mathbf{x}_{0:T-1} \mid \mathbf{x}_T) = \prod_{t=1}^{T} p_\theta(\mathbf{x}_{t-1} \mid \mathbf{x}_t),
\end{equation}  
where \(\theta\) denotes the learnable parameters of the reverse process. The initial state of the reverse process starts from a Gaussian noise distribution \(p_\theta(\mathbf{x}_T) = \mathcal{N}(\mathbf{x}_T; \mathbf{0}, \mathbf{I}_{n \times n})\), where \(\mathbf{I}\) represents the raw input image. Through this iterative reverse process, the latent variable distribution \(p_\theta(\mathbf{x}_T)\) transitions to the target data distribution \(p_\theta(\mathbf{x}_0)\).  

Figure \ref{fig:figure2} illustrates the conditioning of the reverse process on temporal remote sensing (RS) images. This process is defined as follows:  

\begin{equation}
\begin{aligned}
\boldsymbol{\epsilon_{\theta}}(\mathbf{x}_t, I_{\text{pre}}, I_{\text{post}}, t) = 
\text{UNet}(& I_{\text{pre}}, I_{\text{post}},\\ 
            &\text{Cat}(\text{Diff\_Image}, \text{MaskNoise}), t),
\end{aligned}
\end{equation}
where \(\text{Cat}(\cdot)\) represents the concatenation operation. The variables \(I_{\text{pre}}\) and \(I_{\text{post}}\) correspond to the remote sensing images before and after the event, respectively. \(\text{Diff\_Image}\) captures the difference between the two temporal images by Mask Approx Module, while \(\text{MaskNoise}\) introduces noise to the mask during the denoising process.

The estimated noise is iteratively utilized to guide the sampling process at each step, following the equations (\ref{eq1})-(\ref{eq5}). The sampling formula is expressed as:  
\begin{equation}
\mathbf{x}_{t-1} = \sqrt{\frac{1}{\alpha_t}} \mathbf{x}_t - \sqrt{\frac{1 - \alpha_t}{1 - \bar{\alpha}_t}} \boldsymbol{\epsilon}(\mathbf{x}_t, I_{pre}, I_{post}, t) + \tilde{\beta}_t \mathbf{z},
\end{equation}
where \(\mathbf{z} \sim \mathcal{N}(\mathbf{0}, \mathbf{I}_{n \times n})\) is a random vector with elements sampled independently from a standard normal distribution. After performing iterations of this sampling process~\cite{bandara2022revisiting}, change detection (CD) map is generated, starting from Gaussian noise.

\subsubsection{Frequency Noise Filter}
In the UNet denoising network, the presence of multi-scale noisy features, particularly high-dimensional noise, significantly impacts the generation quality of the change map. To address this issue, we introduced the Frequency Noise Filter before producing the final results. The Frequency-Guided Channel Fusion (FGCF) method, as illustrated in Figure \ref{fig:figure3}, is specifically designed for noise suppression in high-dimensional data by leveraging frequency-domain analysis.

$X^\text{Noise} = (F_\text{pre}, F_\text{post}, \text{Diff\_Image}_{k=3,6,9,12})$ represents the noisy features as the input of FGCF, \( F_{\text{pre}} \) and \( F_{\text{post}} \) are the processed features of \( I_{\text{pre}} \) and \( I_{\text{post}} \), respectively, obtained after passing through the feature layers,  \( k \) denotes the multi-scale levels of noisy features. The method starts by transforming each channel of the input tensor into the frequency domain using the Discrete Fourier Transform (DFT), defined as
\begin{equation}
\hat{X}_c = F(X_c^\text{Noise}),
\end{equation}
where \( F \) denotes the Fourier transform and \( \hat{X}_c \in \mathbb{C}^{H \times W} \) is the frequency representation of the \( c \)-th channel. To suppress high-frequency noise, a frequency-guided weighting function \( W_c(f_x, f_y) \) is applied to each frequency component \( (f_x, f_y) \) of \( \hat{X}_c \). The weighting function is defined as

\begin{equation}
W_c(f_x, f_y) = \exp \left( -\alpha \cdot \| F(f_x, f_y) \|_2^2 \right),
\end{equation}
where \( F(f_x, f_y) \) is a learnable frequency filter and \( \alpha \) is a scaling parameter controlling the suppression strength. The filtered frequency representation becomes
\begin{equation}
\hat{X}_c' = W_c \odot \hat{X}_c,
\end{equation}
where \( \odot \) denotes element-wise multiplication. After filtering, the frequency-domain representation is transformed back to the spatial domain using the Inverse Fourier Transform (IFT):
\begin{equation}
X_c' = F^{-1}(\hat{X}_c').
\end{equation}

To aggregate complementary information across channels, a channel attention mechanism is applied, where the channel-wise attention weights \( A_c \) are computed as

\begin{equation}
A_c = \sigma \left( W_a \cdot \text{GAP}(X_c') + b_a \right),
\end{equation}
with \( \text{GAP} \) representing Global Average Pooling, \( W_a \) and \( b_a \) as learnable parameters, and \( \sigma \) as the sigmoid activation function. The final fused feature tensor is obtained by summing the weighted features from all channels:

\begin{equation}
X_{FGCF} = \sum_{c=1}^{C} A_c \cdot X_c'.
\end{equation}

The advantages of FGCF include frequency-domain adaptation, which targets specific frequency bands to suppress noise more effectively than purely spatial-domain methods, and channel fusion, which enhances the network’s ability to preserve essential features while suppressing noise. The method also ensures parameter efficiency, as the learnable components such as the frequency filter \( F \), the channel attention weights \( W_a \), and the bias \( b_a \) are lightweight, introducing minimal computational overhead. These transformations emphasize or suppress specific frequency components, enabling the module to capture and enhance meaningful spatial patterns within the input data.

\begin{table*}[!t]
  \caption{Comparison of Methods' Performance on Multiple Evaluation Metrics on LEVIR-MCI Dataset}
  \centering
  \resizebox{0.8\textwidth}{!}{%
  \begin{tabular}{l|ccccccc}
    \toprule
    \multirow{2}{*}{\textbf{Method}} & \multicolumn{7}{c}{\textbf{Metrics}} \\
    \cmidrule(lr){2-8}
    & \textbf{BLEU-1} & \textbf{BLEU-2} & \textbf{BLEU-3} & \textbf{BLEU-4} 
    & \textbf{METEOR} & \textbf{ROUGE-L} & \textbf{CIDEr-D} \\
    \midrule
    Capt-Rep-Diff\cite{park2019robust} & 72.90 & 61.98 & 53.62 & 47.41 & 34.47 & 65.64 & 110.57 \\
    Capt-Att\cite{park2019robust} & 77.64 & 67.40 & 59.24 & 53.15 & 36.58 & 69.73 & 121.22\\
    Capt-Dual-Att\cite{park2019robust} & 79.51 & 70.57 & 63.23 & 57.46 & 36.56 & 70.69 & 124.42\\
    DUDA\cite{park2019robust} & 81.44 & 72.22 & 64.24 & 57.79 & 37.15 & 71.04 & 124.32 \\
    MCCFormers-S\cite{qiu2021describing} & 82.16 & 72.95 & 65.42 & 59.41 & 38.26 & 72.10 & 128.34 \\
    MCCFormers-D\cite{qiu2021describing} & 80.49 & 71.11 & 63.52 & 57.34 & 38.23 & 71.40 & 126.85 \\
    RSICCformer\cite{liuc} & 84.11 & 75.40 & 68.01 & 61.93 & 38.79 & 73.02 & 131.40 \\
    PSNet\cite{liuc2} & 83.86 & 75.13 & 67.89 & 62.11 & 38.80 & 73.60 & 132.62\\
    Prompt-CC\cite{liuc3} & 83.66 & 75.73 & 69.10 & 63.54 & 38.82 & 73.72 & 136.44\\
    Chg2Cap-S\cite{changs} & 82.41 & 73.10 & 65.29 & 59.02 & 38.71 & 72.47 & 130.88 \\
    Chg2Cap\cite{changs} & 85.14 & 76.91 & {69.86} & 64.09 & 39.83 & 74.62 & 135.41 \\
    Sparse Focus\cite{sundw} & 84.56 & 75.87 & 68.64 & 62.87 & 39.93 & 74.69 & 137.05 \\
    SEN\cite{sen} & 85.10 & 77.05 & 70.01 & 64.09 & 39.59 & 74.57 & 136.02 \\
    DiffusionRSCC\cite{yu2025diffusion} & - & - & - & 60.90 & 37.80 & 71.50 & 125.60\\ 
    \midrule
    Mask Approx Net(ours) & \textbf{85.90} & \textbf{77.12} & \textbf{70.72} & \textbf{64.32} & \textbf{39.91} & \textbf{75.67} & \textbf{137.71}\\
    \bottomrule
  \end{tabular}%
}
  \label{tab1}
\end{table*}

\subsection{Text Decoding Phase}
\textbf{Caption training.}
The process of generating sentences using a transformer decoder begins with the conversion of initial token sequences $\bm{t}$ into word embeddings, represented as $\bm{T}_{\text{embed}}$. This embedding transformation is defined as:
\begin{equation}
\bm{T}_{\text{embed}} = E_{\text{embed}}(\bm{t}) + E_{\text{pos}},
\end{equation}
where $E_{\text{embed}}(\bm{t})$ represents the token embeddings and $E_{\text{pos}}$ is the positional encoding, which captures the sequential order of the tokens. These embeddings serve as the initial input to the transformer decoder. 

Next, $\bm{T}_{\text{embed}}$ is processed via the masked multi-head self-attention (MHA) mechanism, where attention heads are computed as:
\begin{equation}
\bm{Head}_l = \mathrm{Attention}(\bm{T}_{\text{embed}}^{i-1}\bm{W}_l^{Q}, \bm{T}_{\text{embed}}^{i-1}\bm{W}_l^{K}, \bm{T}_{\text{embed}}^{i-1}\bm{W}_l^{V}),
\end{equation}
where the attention function is expressed as,
\begin{equation}
\mathrm{Attention}(\bm{Q}, \bm{K}, \bm{V}) = \mathrm{Softmax}\left(\frac{\bm{QK}^\top}{\sqrt{d}}\right)\bm{V}.
\end{equation}
Here, $\bm{W}_l^{Q}, \bm{W}_l^{K}$, and $\bm{W}_l^{V} \in \mathbb{R}^{d_{\text{embed}} \times d_{\text{embed}/h}}$ are trainable projection matrices for the $l$-th attention head, where $h$ denotes the number of attention heads and $d_{\text{embed}}$ represents the embedding dimension. The outputs from all attention heads are concatenated and linearly projected using $\bm{W}^O \in \mathbb{R}^{d_{\text{embed}} \times d_{\text{embed}}}$, yielding the result of the MHA layer as,
\begin{IEEEeqnarray}{rCl}
\bm{T}_{\text{img}} & = & \mathrm{MHA}(\bm{T}_{\text{embed}}^{i-1}, \bm{T}_{\text{embed}}^{i-1}, \bm{T}_{\text{embed}}^{i-1}) \nonumber\\
& = & \mathrm{Concat}(\bm{Head}_1, \dots, \bm{Head}_h) \cdot \bm{W}^O.
\end{IEEEeqnarray}
Following the MHA layer, the output undergoes further processing through a feed-forward network (FFN). The final output of the decoder layer is obtained by incorporating the input embedding $\bm{T}_{\text{embed}}^{i-1}$ with the output of the FFN via residual connections:
\begin{equation}
\bm{T}_{\text{text}} = \mathrm{FN}(\bm{T}_{\text{img}}) + \bm{T}_{\text{embed}}^{i-1}.
\end{equation}


To generate captions, the decoder output is passed through a linear transformation layer followed by a softmax activation function. This step converts the embedding into probabilities over the vocabulary space, yielding the final caption,
\begin{equation}
\mathrm{Caption}_{T} = \mathrm{Softmax}(\mathrm{LN}(\bm{T}_{\text{text}})),
\end{equation}
where $\mathrm{Caption}_T = [\hat{\bm{t}}_1, \hat{\bm{t}}_2, \dots, \hat{\bm{t}}_n] \in \mathbb{R}^{n \times m}$. Here, $n$ represents the length of the generated caption, $m$ denotes the vocabulary size, and $\hat{\bm{t}}_i$ corresponds to the predicted probability distribution for the $i$-th word.

\textbf{Autoregressive Caption Generation.}
During the validation and testing phases, an autoregressive strategy is employed for caption generation. The decoder begins with a special ``START'' token and iteratively predicts the next token by conditioning on the previously generated tokens and the encoder’s output features. At each step, logits are computed using a linear transformation, and probabilities are derived via softmax. This process continues until the ``END'' token is generated, marking the completion of the caption.

The autoregressive method enables the model to seamlessly incorporate information from the input image along with the tokens generated earlier, thereby guaranteeing the generation of coherent and contextually appropriate captions.

\section{Experimental} \label{Experimental}
\subsection{Datasets}
The LEVIR\_MCI dataset \cite{10591792} is a large-scale benchmark designed for remote sensing image change captioning tasks. It comprises 10,077 pairs of bi-temporal remote sensing images, each with a spatial resolution of 0.5 meters per pixel and a size of 256 $\times$ 256 pixels. These image pairs are sourced from 20 distinct regions in Texas, USA, capturing various urban and suburban developments over time. Each image pair is annotated with five descriptive sentences detailing the changes observed between the two time points, resulting in a total of 50,385 change captions. This dataset facilitates the development and evaluation of models that generate natural language descriptions of changes in remote sensing imagery. 

The WHU-CDC dataset is derived primarily from the WHU-CD \cite{8444434} dataset and includes high-resolution (0.075 m) image pairs with dimensions of 32,507 × 15,354 pixels. These images document the construction of buildings in the Christchurch region following the 6.3 magnitude earthquake in February 2011 in New Zealand. From this source, the WHU-CDC dataset \cite{10740028} provides 7,434 image pairs, each resized to 256 × 256 pixels, and is accompanied by 37,170 sentences describing the changes. The sentences range in length from three to 24 words, with a total vocabulary size of 327 unique words.

\begin{table*}[t!]
  \caption{Comparison of Methods' Performance on Multiple Evaluation Metrics on WHU-CDC Dataset}
  \centering
  \resizebox{0.8\textwidth}{!}{%
  \begin{tabular}{l|ccccccc}
    \toprule
    \multirow{2}{*}{\textbf{Method}} & \multicolumn{7}{c}{\textbf{Metrics}} \\
    \cmidrule(lr){2-8}
    & \textbf{BLEU-1} & \textbf{BLEU-2} & \textbf{BLEU-3} & \textbf{BLEU-4} 
    & \textbf{METEOR} & \textbf{ROUGE-L} & \textbf{CIDEr-D} \\
    \midrule
    Capt-Rep-Diff\cite{park2019robust} & 70.59 & 59.02 & 50.70 & 45.33 & 32.29 & 63.44 & 108.07 \\
    Capt-Att\cite{park2019robust} & 75.62 & 64.68 & 56.89 & 50.62 & 33.97 & 67.17 & 118.33 \\
    Capt-Dual-Att\cite{park2019robust} & 77.15 & 68.47 & 60.81 & 55.42 & 34.22 & 68.04 & 121.45 \\
    DUDA\cite{park2019robust} & 79.04 & 69.53 & 61.57 & 55.64 & 34.29 & 68.98 & 121.85 \\
    MCCFormers-S\cite{qiu2021describing} & 81.12 & 75.04 & 69.95 & 65.34 & 42.11 & {78.52} & {147.09}\\
    MCCFormers-D\cite{qiu2021describing} & 73.29 & 67.88 & 64.03 & 60.96 & 39.69 & 73.67 & 134.92 \\
    RSICCformer\cite{liuc} & 80.05 & 74.24 & 69.61 & 66.54 & 42.65 & 73.91 & 133.44 \\
    PSNet\cite{liuc2} & 81.26 & 73.25 & 65.78 & 60.32 & 36.97 & 71.60 & 130.52\\
    Prompt-CC\cite{liuc3} & 81.12 & 73.96 & 67.22 & 61.45 & 36.99 & 71.88 & 134.50\\
    Chg2Cap-S\cite{changs} & 77.27 &  71.33  &  65.10 &  58.44 & 39.95 &  \textbf{78.67} & \textbf{160.64}\\
    Chg2Cap\cite{changs} & 78.93 & 72.64  & 67.20  & 62.71 & 41.46 & 77.95 & 144.18\\
    Sparse Focus\cite{sundw} & 81.17 & 72.90 & 66.06 & 60.27 & 37.34 & 72.63 & 134.64 \\
    SEN\cite{sen} & 80.60 & 74.64 & 67.69 & 61.97 & 36.76 & 71.70 & 133.57 \\
    DiffusionRSCC\cite{yu2025diffusion} & 75.32 & 70.15 & 66.40 & 63.76 & 40.18 & 73.80 & 127.96 \\
    \midrule
    Mask Approx Net(ours) &\textbf{81.34} & \textbf{75.68} & \textbf{71.16} & \textbf{67.73} & \textbf{43.89} & {75.41} & {135.31} \\
    \bottomrule
  \end{tabular}%
  }
  \label{tab2}
\end{table*}

\subsection{Experimental Setup}
\subsubsection{Evaluation Metrics}
In evaluating natural language generation tasks, several automatic metrics are commonly employed to assess the quality of generated text:
\begin{itemize}
    \item \textbf{BLEU}: \textcolor{deepblue}{A precision-based metric for machine translation that measures n-gram overlap between candidate and reference texts. A brevity penalty is used to penalize short outputs \cite{papineni2002bleu}.}
    
    \item \textbf{ROUGE-L}: \textcolor{deepblue}{Evaluates summary quality based on the longest common subsequence (LCS) between candidate and reference, considering both precision and recall \cite{lin2004rouge}.}

    \item \textbf{METEOR}: \textcolor{deepblue}{Aligns words using exact matches, stemming, and synonyms; computes a harmonic mean of precision and recall with a fragmentation penalty \cite{banerjee2005meteor}.}

    \item \textbf{CIDEr-D}: \textcolor{deepblue}{Designed for image captioning, this metric compares n-gram similarity between candidate and reference descriptions using TF-IDF weighting \cite{vedantam2015cider}.}
\end{itemize}
\subsubsection{Experimental Details}The deep learning methods presented in this study were implemented using the PyTorch framework and executed on NVIDIA A100 GPUs equipped with 80 GB of memory. The training and evaluation processes followed carefully designed parameters. The Adam optimizer \cite{kingma2014adam} was employed with an initial learning rate of 5e-5, while momentum and weight decay were set to 0.99 and 0.0005, respectively. The training process was conducted over approximately 200 epochs with a batch size of 8, ensuring computational efficiency.

\subsection{Quantitative Results}
Table \ref{tab1} summarizes the performance of the proposed Mask Approx Net and competing methods on the LEVIR-MCI dataset across multiple evaluation metrics. Mask Approx Net achieves state-of-the-art results, consistently outperforming all baseline methods.
Notably, Mask Approx Net achieves the highest scores on key metrics such as BLEU-4 (64.32), METEOR (39.91), ROUGE-L (75.67), and CIDEr-D (137.71), demonstrating its superior ability to generate accurate and semantically rich descriptions. Compared to strong baselines such as SEN, Sparse Focus, and Chg2Cap, the proposed method shows consistent improvements, particularly in CIDEr-D and BLEU scores, which are critical for assessing semantic relevance and fluency.

Overall, the results highlight the robustness and effectiveness of Mask Approx Net in capturing detailed semantic and structural information for change captioning, establishing it as the new state-of-the-art on the LEVIR-MCI dataset.

The experimental results on the WHU-CDC dataset, presented in Table \ref{tab2}, highlight the competitive performance of our proposed Mask Approx Net across multiple evaluation metrics. Mask Approx Net achieves the highest scores for BLEU-1 (81.34), BLEU-2 (75.68), BLEU-3 (71.16), BLEU-4 (67.73), and METEOR (43.89), demonstrating its effectiveness in generating accurate and semantically meaningful descriptions.
Compared to MCCFormers-S, which achieves the highest ROUGE-L (78.52) and CIDEr-D (147.09) scores, Mask Approx Net achieves slightly lower results in these metrics but offers a balanced improvement across all BLEU and METEOR scores. This suggests that our method excels in maintaining both precision and semantic relevance, particularly in n-gram-based evaluations.

The performance of Mask Approx Net also surpasses MCCFormers-D and RSICCformer across most metrics, showcasing its robustness and adaptability to the WHU-CDC dataset. These results underscore the strength of our method in handling the challenges of remote sensing image change captioning, achieving state-of-the-art performance in key metrics.
\subsection{Qualitative Visualization}
\begin{figure*} [htb]\centering
    \includegraphics[width=\textwidth]{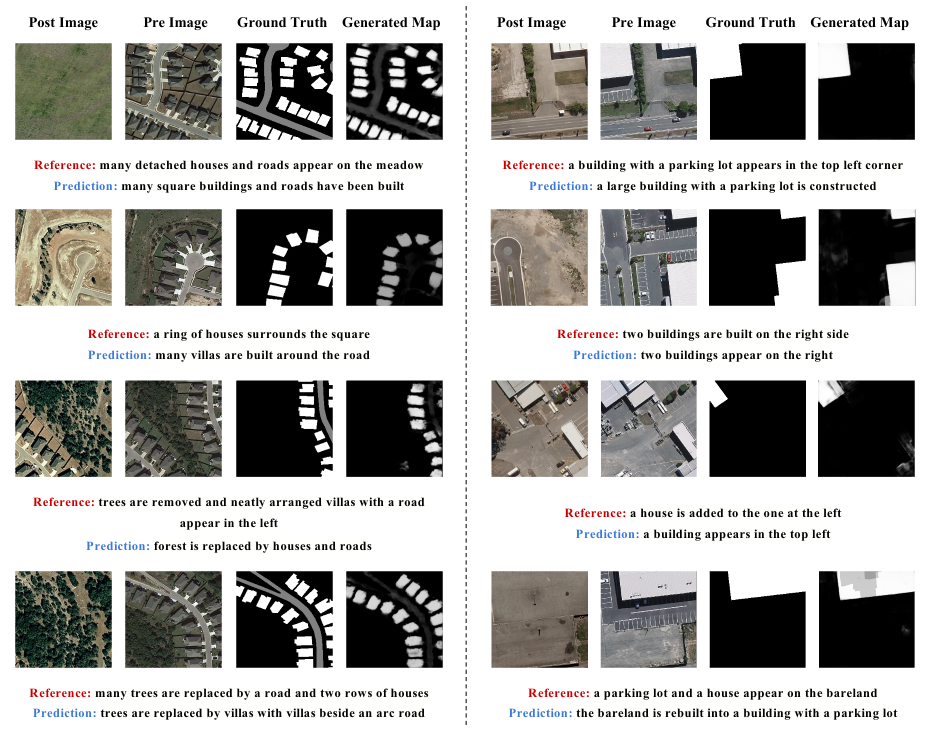}
    \caption{\textcolor{deepblue}{Visualized image and captioning examples generated by Mask Apporx Net on the LEVIR-MCI (THE LEFT) and WHU-CDC (THE RIGHT) dataset}} \label{fig:figure6}
\end{figure*}
The qualitative results in Figure \ref{fig:figure6} demonstrate the effectiveness of the proposed model in capturing semantic changes within remote sensing images. For instance, in scenarios where ``a ring of houses surrounds the square," the model successfully predicts ``many villas are built around the road," indicating a reasonable interpretation of structural changes. The predicted captions effectively capture major transitions, such as the replacement of natural features with urban structures or the emergence of roads and buildings. The generated change maps further validate the model's capacity to align spatial transformations with semantic interpretations, showcasing its potential for practical applications in change detection tasks. Overall, the generated change maps and captions reveal the model's capability to produce contextually relevant and visually coherent descriptions.
\begin{table}[t!]
  \centering
  \caption{\textcolor{deepbluex}{Performance Comparison of Frequency Noise Filter on LEVIR-MCI Dataset}}
  \resizebox{.48\textwidth}{!}{%
    \begin{tabular}{c|c|c|c|c|c|c|c}
    \toprule
    \textbf{FGCF} & \textbf{BLEU-1} & \textbf{BLEU-2} & \textbf{BLEU-3} & \textbf{BLEU-4} & \textbf{METEOR} & \textbf{ROUGE-L} & \textbf{CIDEr-D} \\
    \midrule
    - & 85.63 & 76.79 & 70.41 & 63.96 & 39.70 & 75.53 & 137.24 \\
    \checkmark & \textbf{85.90} & \textbf{77.12} & \textbf{70.72} & \textbf{64.32} & \textbf{39.91} & \textbf{75.67} & \textbf{137.71} \\
    \bottomrule
    \end{tabular}%
  }
  \label{tab:addlabel3}%
\end{table}
The qualitative results presented in Figure \ref{fig:figure6} highlight the model's ability to effectively identify and describe semantic changes in remote sensing images. For instance, the predictions align closely with the ground truth descriptions, such as recognizing the construction of ``a building with a parking lot in the top left corner" or the transformation of ``bareland into a building with a parking lot." These examples underscore the model's capacity to capture both spatial and contextual details of changes, such as the addition of structures or modifications to existing landscapes. The generated change maps complement the captions by visually reinforcing the predicted transformations, demonstrating consistency between visual and textual outputs. This cohesive alignment between predicted captions and change maps illustrates the model's potential to support real-world applications in monitoring urban development and environmental changes.

\begin{table}[!t]
  \centering
  \caption{Comparison of Model Parameters and Computational Cost (FLOPs) Across Different Methods with $256\times 256$ image size}
    \begin{tabular}{cccc}
    \toprule
    \textbf{ID} & \textbf{Method} & \textbf{Paramters(M)} & \textbf{FLOPs} \\
    \midrule
    1     & Capt-Rep-Diff & 45.6 & 19.8 G \\
    2     & Capt-Att & 46.1 & 19.9 G \\
    3     & Capt-Dual-Att & 48.1 & 20.0 G \\
    4     & DUDA  & 52.8 & 20.3 G \\
    5     & MCCFormer-S  & 135.0 & 25.1 G \\
    6     & MCCFormer-D  & 135.0 & 25.1 G \\
    7     & RSICCFormer & 145.3 & 27.1 G \\
    8     & PSNet & 231.5 & 13.8 G \\
    9     & Prompt-CC  & 196.3 & 19.9 G \\
    10    & Chg2Cap & 286.5 & 232.1 G \\
    11    & Sparse Focus & 117.5 & 21.5 G \\
    12    & SEN   & 39.9 & 10.9 G \\
    13    & DiffusionRSCC & 103.0 & 15.2 T \\
    \midrule
    14    & Mask Apporx Net(ours) & 202.5 & 12.3 T \\
    \bottomrule
    \end{tabular}%
  \label{tab:addlabel}%
\end{table}%

\subsection{Ablation Study}
\textcolor{deepbluex}{The ablation results in Table \ref{tab:addlabel3} clearly demonstrate the effectiveness of the frequency noise filter. Compared to the baseline without the filter, the model with the filter achieves consistent improvements across all evaluation metrics. Specifically, the frequency noise filter increases BLEU-1 to BLEU-4 scores by 0.27, 0.33, 0.31, and 0.36, respectively. In addition, gains are also observed on METEOR (from 39.70 to 39.91), ROUGE-L (from 75.53 to 75.67), and CIDEr-D (from 137.24 to 137.71), further confirming the positive impact of the proposed filter on captioning performance.}

\begin{figure}[!t] \centering
    \includegraphics[width=0.5\textwidth]{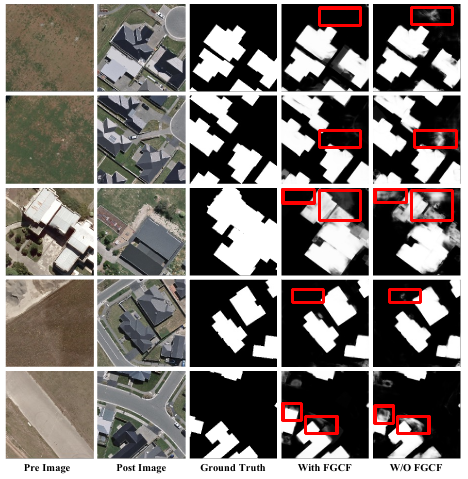}
    \caption{\textcolor{deepblue}{Comparison of mask approximation results with and without the proposed FGCF method.}} 
    \label{fig:figure7}
\end{figure}

In this ablation experiment, we also visualize the effectiveness of the proposed FGCF method in improving mask approximation performance. The Figure \ref{fig:figure7} compares results from the baseline configuration (``Without FGCF") and the enhanced model (``With FGCF"), alongside the ground truth for reference. It's observed that incorporating FGCF significantly improves the accuracy of the mask approximations, with results aligning more closely to the ground truth. The FGCF-enhanced model effectively suppresses noise and preserves critical structural details in the masks. These results confirm the effectiveness of FGCF in addressing challenges related to high-frequency noise suppression and spatial consistency, underscoring its critical role in enhancing the overall model performance.

\textcolor{deepblue}{We compared model parameters and computational cost across different methods on A100 GPU. As shown in Table \ref{tab:addlabel} our Mask Approx Net exhibits a relatively larger parameter count and computational cost compared to several existing methods. This is largely attributable to the use of diffusion models, which involve iterative sampling procedures that enhance the model's expressiveness. Despite this, our approach consistently achieves state-of-the-art performance across multiple datasets, demonstrating the effectiveness of this design choice.}

\textcolor{deepblue}{Notably, some competing methods with fewer parameters or lower FLOPs achieve strong results; however, our method’s incorporation of diffusion modeling provides enhanced accuracy and robustness, reflecting a favorable trade-off in practical applications. Furthermore, the strong theoretical foundation of diffusion models not only supports the empirical gains observed here but also suggests promising directions for future optimization to reduce computational demands while maintaining or even improving performance.}

\textcolor{deepbluex}{We conducted an ablation study on the designed Mask Approx Module to demonstrate its simplicity and effectiveness. As shown in the Table \ref{tab:10}, ResNet50 achieves the best balance between performance and complexity yielding a BLEU-4 score of 57.47 and a CIDEr-D score of 129.16. So we choose ResNet50 as the backbone network. Secondly, introducing the multi scale attention feature fusion (MSAFF) module can effectively fuse image features, raising BLEU-4 to 62.15 and CIDEr-D to 135.16. Finally, when we further incorporate the convolutional enhancement (CE) branch which employs a $3 \times 3$ convolution to emphasize fine details and generate convolutional change encodings into the Baseline $+$ MSAFF setup, we obtain our best overall performance with BLEU-4 reaching 64.32 and CIDEr-D increasing to 137.71.}
\begin{table*}[t!]
  \centering
  \caption{\textcolor{deepbluex}{Ablation study between model complexity and captioning performance with backbone selection and proposed modules on LEVIR-MCI Dataset}}
  \resizebox{.9\textwidth}{!}{
    \begin{tabular}{l|ccccccc|c}
    \toprule
    \multicolumn{1}{c|}{\textbf{Method}} & \textbf{BELU-1} & \textbf{BELU-2} & \textbf{BELU-3} & \textbf{BELU-4} & \textbf{METEOR} & \textbf{ROUGE-L} & \textbf{CIDEr-D} & \textbf{Params(M)} \\
    \midrule
    ResNet34 & 70.12 & 59.27 & 52.34 & 46.15 & 32.88 & 63.89 & 109.08 & 21.2 \\
    ResNet50 (Backbone) & 80.63 & 71.75 & 64.14 & 57.47 & 36.51 & 70.74 & 129.16 & 24.3 \\
    ResNet101 & 81.12 & 71.47 & 63.53 & 57.73 & 36.88 & 70.82 & 129.02 & 42.5 \\
    Backbone + MSAFF & 83.12 & 74.83 & 68.07 & 62.15 & 37.65 & 73.03 & 135.16 & 79.3 \\
    \midrule
    Backbone + MSAFF + CE (ours) & \textbf{85.90} & \textbf{77.12} & \textbf{70.72} & \textbf{64.32} & \textbf{39.91} & \textbf{75.67} & \textbf{137.71} & 81.7 \\
    \bottomrule
    \end{tabular}%
}
  \label{tab:10}%
\end{table*}%

\subsection{Limitation and Future Work}
For the first time, we applied the diffusion model approach to the task of remote sensing image change captioning. Through this study, we summarize the existing challenges and future directions as follows: 1. The training time of diffusion models, as well as their slow convergence and inference speed during training, remains a fundamental issue. 2. As an application-oriented research field, remote sensing image processing will increasingly focus on the model's size, generalization ability, and robustness. 3. The semantic and logical alignment between images and descriptive text remains a key focus for future research.
\section{Conclusion} \label{Conclusion}
In this paper, we presented a novel approach to remote sensing image change captioning by integrating diffusion models into the change captioning process. Our method shifts the focus from traditional feature learning paradigms to a data distribution learning perspective, addressing the limitations of existing CNN-based techniques. By incorporating a multi-scale change captioning module and refining the output features through a diffusion model, we have shown significant improvements in the adaptability of change captioning across different datasets. Furthermore, the introduction of the frequency-guided complex filter module ensures that high-frequency noise is effectively managed during the diffusion process, thereby preserving model performance.

The experimental results on multiple remote sensing change captioning description datasets validate the effectiveness of our proposed method, demonstrating superior performance compared to existing approaches. Our framework not only improves captioning accuracy but also enhances interpretability, contributing to a more comprehensive understanding of changes in remote sensing imagery. Future work will explore further optimizations of the diffusion process and extend the applicability of our approach to other multimodal tasks in remote sensing, ensuring even broader impact and usability in real-world applications.
\bibliographystyle{IEEEtran}
\bibliography{ref}
\vfill

\end{document}